\documentclass[11pt,twocolumn]{article}

\usepackage[T1]{fontenc}
\usepackage[utf8]{inputenc}
\usepackage[margin=0.85in]{geometry}
\usepackage{amsmath}
\usepackage[american]{babel}
\usepackage{amsfonts}
\usepackage{setspace} 
\usepackage{graphicx}
\usepackage{booktabs}
\usepackage{threeparttable}
\usepackage{hyperref}
\hypersetup{hidelinks}
\usepackage{color}
\usepackage{listings}
\usepackage{xcolor}
\usepackage{graphicx}   
\usepackage{hyperref}   

\newcommand{\orcid}[1]{\href{https://orcid.org/#1}{\includegraphics[width=9pt]{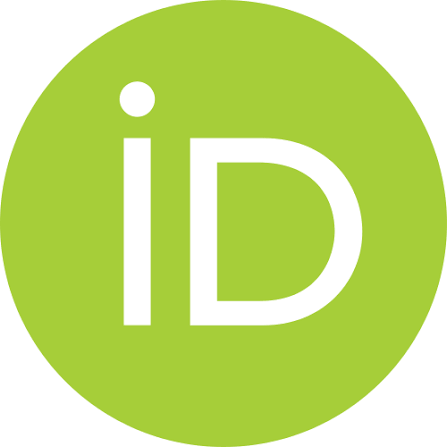}}}

\usepackage[round,authoryear]{natbib}
\title{Automated item evaluation: Predicting item acceptance and rejection using LLM-generated critiques}

\author{
  {\large Hotaka Maeda\,\orcid{0009-0000-9498-786X}\textsuperscript{1} \quad
          Yikai Lu\,\orcid{0000-0003-4410-2589}\textsuperscript{2}}\\[0.5ex]
  \textsuperscript{1}Smarter Balanced, University of California-Santa Cruz\\
  \textsuperscript{2}Department of Educational Psychology, University of Minnesota-Twin Cities
  \thanks{%
    Correspondence concerning this article should be addressed to Hotaka Maeda, Smarter Balanced, University of California-Santa Cruz, 1156 High St, Santa Cruz, CA 95064. E-mail: \href{mailto:hotaka.maeda@smarterbalanced.org}{hotaka.maeda@smarterbalanced.org}%
  }
}

\date{}

\newcommand{\parencite}{\citep}
\newcommand{\textcite}{\citet}

\newcommand{\figurenote}[1]{\par\smallskip\footnotesize\emph{Note.} #1\par}

\newcommand{\tablenote}[1]{\par\smallskip\footnotesize\raggedright\emph{Note.} #1\par}

\begin{document}

\maketitle

\begin{abstract}
\textit{Automated item evaluation} (AIE) refers to the use of computational methods to assess item quality without requiring manual expert review or field testing of the items under evaluation. We aimed to build a near-comprehensive AIE model by predicting item acceptance and rejection from item text using historical rejection data from a large-scale standardized testing program. The dataset contained 52,759 English language arts (ELA) and mathematics items with 34\% permanently rejected from future operational use. Rejection reasons included poor psychometric properties, content issues, bias and sensitivity concerns, and non-content issues. We fine-tuned a DeBERTaV3-large classifier on raw item text, a second DeBERTa classifier on Qwen3-generated item critiques, and a fusion model combining representations from both. The fusion model achieved the strongest overall performance (Accuracy $= .75$, F1 $= .64$, AUC$= .80$, Sensitivity $= .64$, Specificity $= .81$). Prediction for math (F1~$= .73$, AUC~$= .86$) was considerably more accurate than ELA (F1~$= .51$, AUC~$= .72$). Lowering the decision threshold from .5 to .25 raised average sensitivity for ELA and math to .88 and .91, while reducing specificity to .31 and .56, respectively, which may be preferable in automated item generation contexts where generating items is cheaper than evaluating them. Incorporating item critiques alongside raw item text improved performance across most rejection reasons. The model assigned higher rejection probabilities to more difficult items. However, the fusion model struggled to identify items flagged for bias, sensitivity, fairness, or accessibility, especially for ELA. These findings suggest that text-based AIE is feasible in some areas and may offer a practical tool for reducing the burden of manual review and field testing, while also underscoring the importance of human review for items with fairness concerns.
 
\end{abstract}

\noindent\textbf{Keywords:} natural language processing, Transformers, artificial intelligence, item difficulty prediction, item response theory


\section{Introduction}

Evaluating the quality of new items is expensive and time-consuming, requiring content review and field testing. This issue has grown since the emergence of automated item generation (AIG) using large language models (LLMs). AIG review papers consistently urge for methods that can quickly filter poor quality items \parencite{oluoke_inpress, circi2023automatic, falcao2022feasibility, tan2025review}. Relevant literature has focused on using LLMs, natural language processing (NLP), or machine learning to predict individual qualities of items, including item difficulty \parencite{alkhuzaey2023text, benedetto2023survey}, content alignment with the assessment blueprint \parencite{fu2025textbasedapproachesitemalignment}, or differential item functioning \parencite{maeda_finding_2025}. However, prior work has not addressed these concerns as a unified class of issues related to item quality. To capture this broader problem space, we use the term \textit{automated item evaluation} \parencite[AIE; ][]{yuaie}, extending its prior use from a much narrower scope. In this study, AIE refers to the use of computational methods to automatically assess item quality without requiring manual expert review or field testing of the items under evaluation.



The difficulty of AIE stems from the fact that there are countless reasons that items can be unsuited for operational use. The Standards for Educational and Psychological Testing \parencite{standards} articulate the broader requirements that define a valid, reliable, and fair item. Also, \textcite{haladyna1989taxonomy} cataloged 43 item-writing rules covering stem construction, option formatting, correct answer selection, and distractor quality. Training and deploying a separate prediction model for each rule is unrealistic because model training requires large sample sizes and substantial computational resources. Moreover, treating rules separately fails to leverage the fact that items may exhibit multiple issues simultaneously and does not directly support the goal of rejecting the most undesirable items.


Therefore, a comprehensive AIE approach could be desirable. One possible approach is to use an existing item bank by treating operational items as ``accepted'' and items permanently retired due to irreparable issues as ``rejected'', then training a model to predict that outcome from item content alone. Accepted items had to have gone through rigorous content expert review and field testing, so this single label becomes an all-encompassing measure of item quality. Although this approach requires a large established item bank, it may be the single most comprehensive AIE method for both AIG and human-written items. 

To our knowledge, no studies have predicted item acceptance and rejection as an AIE method. The purpose of this study is to train a transformer language model to classify items as accepted or rejected for operational use, using item status data from a large-scale standardized testing program. To enhance predictive power, we augment the data using item critiques generated by an LLM. To provide insight into model behavior beyond overall classification performance, we use historical item developer comments to evaluate classifier sensitivity by rejection reason. We also include a sentiment analysis to further understand how LLM critique sentiment is related to prediction model behavior.

\section{Related Works}
 
\subsection{Item Evaluation}

Item evaluation is an important step during test development \parencite{standards,haladyna1989taxonomy}. \textcite{Gorgun2025} summarized existing item evaluation approaches into three categories: metric-based evaluations, post-hoc analysis, and human evaluations. Metric-based approaches typically focus on comparing items with a set of reference items. This includes NLP-based techniques, such as word count, TF-IDF \parencite{benedetto2020r2de}, cosine similarity for identifying enemy items, and measures of readability and text complexity \parencite{yuaie,amini2025promptingstrategieslanguagemodelbased}. As a more sophisticated example, \textcite{pelanek2022towards} proposed a system that uses an outlier detection mechanism to identify computationally or statistically deviant items, thereby flagging potentially \textit{bad} items. However, despite their convenience and potential efficiency, metric-based approaches are limited by their reliance on reference items. This dependence can significantly restrict the content domain of generated items, as valid items may be incorrectly penalized simply because they \textit{appear} different from reference items at the level of linguistic features \parencite{Gorgun2025}.

Alternatively, post-hoc analysis approaches typically involve field testing to collect real item response data from a representative sample of examinees to assess item quality. Specifically, the Standards recommend meeting psychometric specifications related to item difficulty, item discrimination, and the absence of differential item functioning \parencite{standards}. However, this approach can be costly and time-consuming, as it requires a substantial sample size to obtain reliable statistical estimates. Finally, human evaluators can identify problems in items that may be difficult to detect using computational or statistical approaches. However, human evaluation can be subjective and time-consuming, especially when many items must be reviewed. Finding suitable experts for the evaluation process is also challenging.

In summary, metric-based approaches are automatic but limited in scope. Post-hoc analyses and human evaluations can be high-quality but are resource-intensive. However, we argue in the next section that AIE can be designed to take the advantages of all three, while bypassing some of their limitations.

\subsection{``Automated'' Item Evaluation}

The term ``automated item evaluation'' or a similar term has been used occasionally, typically in the AIG literature \parencite{Gorgun2025,electronics14112279,yuaie,shin2024automated,wang2025using}. Despite its sparse usage and variation in scope in the past literature, definitions of AIE appear to share a common theme: the use of computational methods to automatically assess some facets of item quality without requiring manual expert review or field testing of the items under evaluation. Among the three categories described by \textcite{Gorgun2025}, only the metric-based approach meets the criteria for AIE, as it eliminates the need to rely on field testing or human evaluators.

That being said, the results obtained from post-hoc analyses or human evaluations could also serve as labels for training machine learning models or LLMs to build AIE prediction models. Predicting human labels was proposed by \textcite{Gorgun2025}, where an LLM was trained to identify low-quality AIG items. However, such an approach is limited because it provides only a coarse-grained evaluation based on human review labels, classifying items simply as \textit{good} or \textit{bad}. \textcite{pelanek2022towards} argues that such classifications may be a fundamentally ill-defined problem, as items that warrant the attention of content creators depend on the specific context and are at least partly subjective. Furthermore, items may be problematic for many different reasons, including linguistic or psychometric characteristics that sometimes cannot be captured by human evaluators and vice versa. On the other hand, the prediction of statistical/psychometric properties using past post-hoc analyses data has received considerable attention. The most commonly examined aspects are statistical item properties, such as item difficulty prediction \parencite{alkhuzaey2023text, benedetto2023survey} and item discrimination prediction \citep[e.g.,][]{multimodal2025, han2025leveraging, yaneva-etal-2020-predicting}. Less commonly examined areas include prediction of differential item functioning \parencite{maeda_finding_2025}. Although linguistic information was used as model input, the evaluation outcomes were limited to psychometric characteristics rather than directly assessing linguistic or semantic dimensions of item quality that could be evaluated through other approaches.

Furthermore, LLM evaluators could be considered a fourth category of item evaluation. LLMs can mimic human evaluators but are automated, and therefore classified as an AIE method. This approach differs from fine-tuning LLMs because it does not require labeled training data and can be implemented in a zero-shot setting. Notably, LLM-based evaluation has enabled the automated assessment of item-content alignment with assessment blueprints, an area of AIE that has received comparatively limited attention \parencite{fu2025textbasedapproachesitemalignment}. This suggests that LLM-based evaluation could shed light on aspects of items that were difficult to evaluate automatically.

The major limitation of the metric-based and psychometric-prediction approaches is that they often only focus on one aspect of item quality at a time. In practice, however, different dimensions of item quality may be interrelated. For example, a psychometrically difficult mathematics item may also contain difficult words, and such relationships cannot be fully leveraged when these issues are evaluated separately. Furthermore, relying solely on NLP or psychometric evaluations may overlook problems that can only be identified through human judgment, since these evaluations often focus only on linguistic features. For these reasons, it is preferable to construct a model that can evaluate multiple dimensions of item quality simultaneously. Our comprehensive approach can explicitly account for these different aspects of item quality by using a dataset that includes rejected items with a broad range of documented reasons for rejection. Rather than relying solely on human evaluators, psychometric analyses, or linguistic features, this approach fine-tunes a Transformer language model to distinguish between acceptable and unacceptable items while implicitly learning from multiple sources of item-quality evidence by taking into account the content of items.

\subsection{Transformer language models for text classification}

The rise of Transformer language models increased demand for AIE research by making large-scale AIG feasible, while simultaneously expanding the computational tools useful for AIE. Traditionally, prediction from text in the assessment context has relied on expert judgment \parencite{wauters2012item}, syntactic features such as word count and term frequency \parencite{benedetto2020r2de}, or semantic features like word embeddings \parencite{hsu2018automated}. More recently, Transformer-based language models have substantially improved prediction accuracy \parencite{li2025item,han2025leveraging,maeda2024field}. Introduced by \textcite{vaswani2017attentionneed}, the Transformer architecture replaced recurrent networks with parallel attention mechanisms, enabling efficient training and scalable modeling of long-range dependencies. Models such as BERT \parencite{Devlin2019} are pre-trained on large corpora and fine-tuned for downstream tasks. Text is tokenized into embeddings and passed through encoder layers to produce contextually enriched representations.

In this study, we use DeBERTaV3-large \parencite{he2021debertav3improvingdebertausing}. DeBERTa extends BERT \parencite{Devlin2019} and RoBERTa \parencite{liu2019roberta}. DeBERTa outperforms earlier models by disentangling content and positional information within its attention mechanism, where each token is represented by separate content and position vectors, allowing the model to capture relationships between words more effectively. The V3 variant further incorporates replaced token detection, yielding a 304-million-parameter model with strong benchmark performance that even outperforms the newer ModernBERT in natural language understanding \parencite{warner2024smarter}.

\subsection{Using LLM Critiques for AIE}

Recently, language models have been used to extract reasoning or rationales to augment a task-specific model \parencite{henrichsen2025twostagereasoninginfusedlearningimproving, hsieh2023distillingstepbystepoutperforminglarger, scarlatos2025smart}. For an example in the assessment context, GPT-4o model \parencite{openai2024gpt4ocard} was used to generate the reasoning steps required to reach each multiple choice item option, and the responses were inputted into the longformer encoder language model to predict item difficulty \parencite{feng2025reasoningsamplingaugmentedmcqdifficulty}. In another example, LLMs were given the role of content experts to make absolute and pairwise item comparisons to estimate item difficulty \parencite{kolesnikova2026estimating}. 

We use a similar data augmentation technique in this study using Qwen3 \parencite{qwen3technicalreport}. The Qwen3 model family includes a series of compact dense variants spanning 0.6B, 1.7B, 4B, and 8B parameters, all released publicly under the Apache~2.0 license. These models follow the broader trend of developing capable small language models (SLMs) suited to resource-constrained settings, where low memory footprint and fast inference are primary constraints \parencite{lu2025smalllanguagemodelssurvey}. A key capability distinguishing these models from earlier SLMs is the integration of a \textit{thinking mode} and a \textit{non-thinking mode} within a single model. Rather than maintaining separate chat and reasoning model variants, users can toggle between slow, multi-step reasoning and fast, context-driven response generation at inference time.

Empirical results reported in \textcite{qwen3technicalreport} indicate that the Qwen3-8B, 4B, and 1.7B base models each outperform Qwen2.5 models of the next larger size class on standard benchmarks, reflecting substantial efficiency gains from the distillation-centric training approach.

\section{Methods}  

\subsection{Item Data}

Included in the study were 52,759 items designed for English language arts (ELA) and mathematics state summative assessments for grades 3--11. Of these, 34\% have been rejected, defined as permanently removed from future operational use (30\% for ELA, 39\% for math). Rejection can occur at any stage of item development, before or after field testing, or after some operational use. Rejection reasons span a range of categories including content and standards alignment issues, psychometric reasons, accessibility and sensitivity flags, or issues unrelated to the item content, such as data integrity problems (see~\ref{sec:rejection_reasons} for full category descriptions).

Items that had not been rejected and had not yet been field tested were excluded from the study, as their operational suitability remains unknown. Items rejected prior to field testing were retained, as their unsuitability for operations is established regardless of psychometric properties. In total, 14\% of items in the study are items rejected prior to field testing. The remaining 86\% of items were field tested among grade-matched students across multiple states in the United States. 

To prepare each item for modeling, item text was concatenated with a \texttt{\textbackslash n} separator between prompts and any available answer options. Both selection-based (e.g., multiple choice, multiple select) and constructed response (e.g., short answer, extended response) items were included in the study. Any reading or listening passages were excluded from the item text data due to their extensive lengths. Any images, figures, or tables in the items were also excluded. Metadata were exluded as well, which included the scoring rubric and correct answer keys. Data were partitioned randomly into approximately 80\% training, 10\% validation, and 10\% test data. Item groups with multiple items associated with the same stimulus were always included in the same data group. 

\subsection{Prediction Approaches}

Five prediction approaches were compared. We used a single NVIDIA A10G Tensor Core 24GB graphics processor for all model inferences and training.

\subsubsection{Zero-Shot Baseline}
Qwen3-0.6B \parencite{qwen3technicalreport} was prompted to classify each item as ``accept'' or ``reject'' directly, without any task-specific fine-tuning (see~\ref{sec:zero_prompt} for prompt). The subject, grade level, and the aligned common core standards for the item were included in the prompt. 

\subsubsection{Raw Item Text Fine-Tuned Model}
Item text was tokenized and passed into DeBERTaV3-large \parencite{he2021debertav3improvingdebertausing}, fine-tuned as a binary classifier using Low-Rank Adaptation (LoRA; \citealt{hu2022lora}; LoRA rank = 32, LoRA Alpha = 64, LoRA dropout = 0.05, dropout = 0.1, batch size = 8, max token length = 512, weight decay = .01, learning rate = 5e-5, epochs = 3). LoRA is a parameter-efficient fine-tuning method that inserts trainable low-rank matrices into selected layers while keeping the pretrained weights frozen, substantially reducing memory and computational costs. We added a binary classification head with a sigmoid function to convert logits into rejection probabilities, followed by cross-entropy loss (CEL) to minimize the distance between predicted and true labels.

There were approximately double the number of accepted items compared to rejected items. This class imbalance was addressed by incorporating class weights into the binary CEL. Weights were computed as inversely proportional to class frequency, such that the minority class received proportionally greater influence on the gradient updates during training. This approach preserves the full training sample while discouraging the model from defaulting to the majority class, effectively rebalancing the loss contribution of each class without altering the composition of the training data. By increasing the relative penalty for misclassifying rejected items, class weighting was expected to make the model more attentive to rejected items. We retained the state with the lowest validation CEL out of all epochs. 

\subsubsection{Item Critique Fine-Tuned Model}
To augment the raw item text, Qwen3-0.6B \parencite{qwen3technicalreport}  with thinking mode enabled was prompted to generate a two sentence critique or praise of each item's quality. The subject, grade level, and the aligned common core standards for the item were included in the prompt. See ~\ref{sec:critique_prompt} for the prompt and ~\ref{sec:example_critiques} for example critiques. The resulting critique text was then used as input to DeBERTaV3-large in place of the raw item text, with the same fine-tuning procedure as the text-only model (LoRA rank = 50, LoRA Alpha = 100, LoRA dropout = 0.05, dropout = 0.1, batch size = 8, max token length = 512, weight decay = .01, learning rate = 5e-5, epochs = 3). 
 
\subsubsection{Text + Critique Fusion Model}
\begin{figure*}[!t] 
    \centering
    \caption{Automated item evaluation model using raw text + LLM Critique fusion.}
    \includegraphics[width=\textwidth]{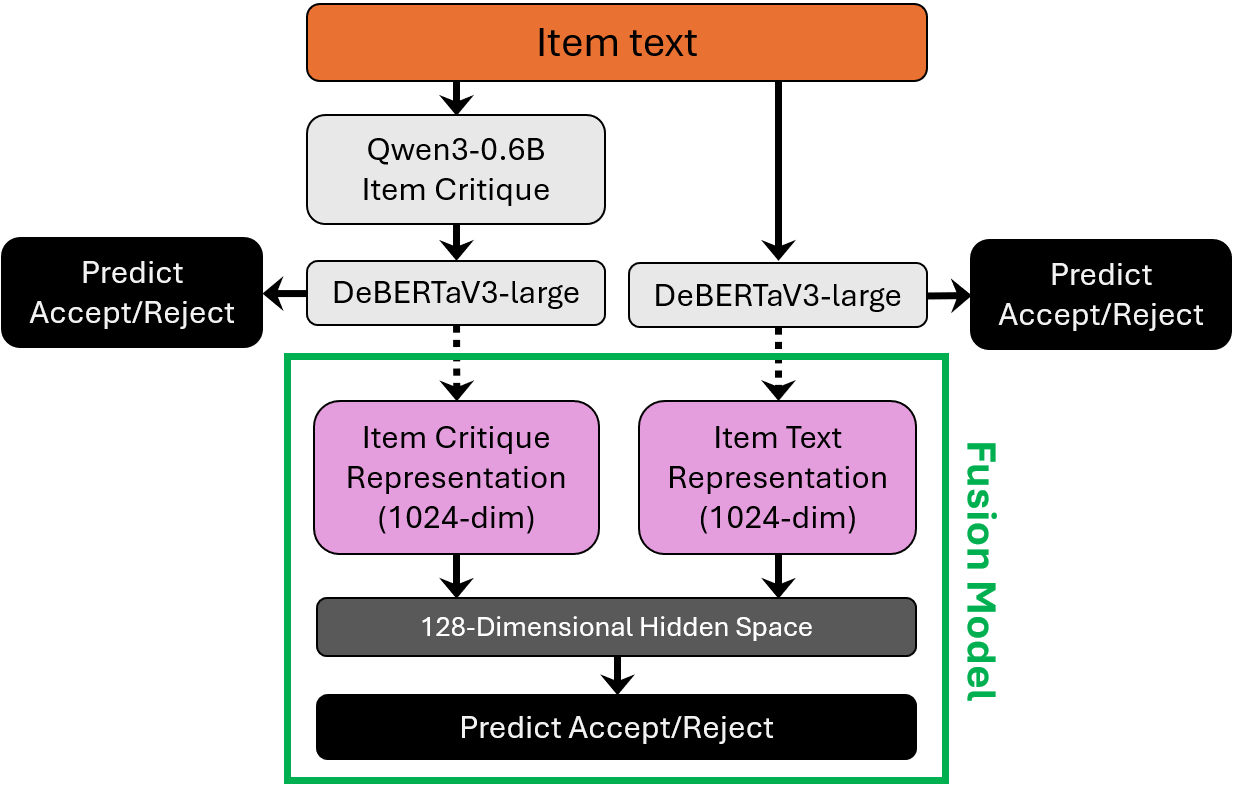}
    \figurenote{Item text is processed directly through the first Deberta. Qwen is used to generate item critiques, then entered into a separate Deberta. Two DeBERTas are combined in the final fusion model.}
    \label{fig:figure_model}    
\end{figure*}

The fusion model combines representations from both DeBERTaV3-large models described above. The 1024-dimensional output of each model's final layer was extracted and concatenated. The DeBERTa weights were frozen, and only the fusion layers were trained. We added a 128-dimensional intermediate layer followed by a binary classification head with a sigmoid function and a weighted CEL (see Figure ~\ref{fig:figure_model}).

\subsubsection{Stand-Alone Subject Models}
Given that ELA and math are very distinct subjects, we included an alternative approach of building separate models for these subjects. All raw text-only, critique-only, and fusion models were re-trained using this approach.


\subsection{Evaluation}

The area under the ROC curve (AUC) was used as a cutoff threshold-independent measure of the model's overall discriminative ability. Classification thresholds were fixed to .50 by default. Using this threshold, we report accuracy (proportion of correctly classified items), precision (proportion of flagged items that are truly rejected), sensitivity (proportion of rejected items correctly flagged), specificity (proportion of viable items correctly passed), and F1 score (harmonic mean of precision and sensitivity). Similar to outlier detection, this paper treats rejected cases as positive cases for evaluation purposes.


\subsubsection{Rejection Reason Classification}
Item developers often left comments when moving an item to a permanent rejection status. Comment quality varied widely, ranging from absent or uninformative (e.g., ``Moving to rejected'') to detailed explanations. Multiple comments for the same item were concatenated and treated as a single record. Using a combination of Claude Sonnet 4.6 \parencite{anthropic2025claude} and manual fixes, comments from rejected items were cleaned and classified into 10 rejection reason categories: content, psychometric, data review, bias, abandoned,  incomplete, passage, scoring, non-content, and no data (see~\ref{sec:rejection_reasons} for definitions). These labels were used to assess how well the best-performing model detected items across each rejection reason. The rejection reason data were only used for test sets. Only 19\% of rejected test items had usable rejection reason data. Other items were classified in the ``no data'' category because the comments were missing (80\%) or were ambiguous (1\%).

\subsection{Rejection Rate by Difficulty and Discrimination}

We examined the Spearman correlations between the fusion model predictions and several item difficulty and discrimination metrics, including proportion correct, item-total correlation, and item response theory (IRT) parameters. All items were calibrated using the two-parameter logistic (2PL) model \parencite{birnbaum1968latent} for dichotomous items or the generalized partial credit model (GPCM; \citealt{muraki1992generalized}) for polytomous items. For GPCM items, the item difficulty parameter was centered to represent average difficulty of the thresholds. Therefore, both item discrimination ($a$) and difficulty ($b$) parameters were comparable between 2PL and GPCM. The data spanned two subjects and multiple grades, but we standardized $\theta$ to have a mean of zero and standard deviation of one within each subject-grade combination, effectively removing the vertical scale for this analysis. Therefore, including proportion correct, item-total correlation, all item statistics are relative to the subject and grade. For example, a low proportion correct for an ELA grade 3 item is difficult for grade 3 students. 

\subsection{Sentiment Analysis of Critiques}
To examine if sentiment can partly explain how the Qwen3 critiques improved prediction, we conducted a sentiment analysis. We computed sentiment using SiEBERT \parencite{hartmann2023}, a RoBERTa-large model fine-tuned for binary sentiment classification across 15 heterogeneous English text sources, chosen for its cross-domain generalization relative to models trained on a single text type. For each critique, we extracted the difference between the positive and negative class logits, yielding a continuous log-odds sentiment score in which higher values indicate more positive sentiment. We then computed Spearman correlations between critique sentiment and predicted rejection probability of the text-only model $P_{text}$, fusion model $P_{fusion}$ and the difference $P\Delta = P_{fusion}-P_{text}$.

\section{Results}  

All results are based on test data. Model performance is summarized in Table~\ref{tab:model_comparison}. Based on F1 score, the zero-shot approach was clearly the worst performer (F1~$=.23$). The fusion model consistently outperformed both the raw text-only and critique-only models (Accuracy~$= .75$, F1~$= .64$, AUC~$= .80$, precision $=.63$, sensitivity~$= .64$, specificity~$= .81$). In all cases, prediction for math was considerably more accurate than ELA. The stand-alone fusion models separated by subject (ELA and math) did not meaningfully change performance metrics. Note that a fusion model with non-weighted CEL had essentially the same performance and was not worth reporting separately (Accuracy~$= .75$, F1~$= .63$, AUC~$= .80$, precision~$=.62$, sensitivity~$= .65$, specificity~$= .79$). However, the weighted model offers an interpretive advantage: because class weighting shifts the optimal decision boundary near 0.5, the classification threshold aligns with the intuitive probability midpoint. Subsequent analyses focus on evaluating the fusion model performance in detail. 

\begin{table*}[!t] 
    \centering
    \begin{threeparttable}
        \caption{\label{tab:model_comparison}Classification Performance by Model and Subject} 
        \footnotesize
        \begin{tabular}[t]{lccccccccc}
            \toprule
            Model & p & AUC & Accuracy & Precision & Sensitivity & Specificity & F1\\
            \midrule
            Zero-shot                            & .17 & .50 & .61 & .35 & .17 & .83 & .23\\
            Raw text-only                   & .31 & .76 & .73 & .61 & .56 & .82 & .59\\  
            \hspace{1em}ELA                 & .21 & .69 & .72 & .55 & .39 & .86 & .46 \\
            \hspace{1em}Math                & .42 & .80 & .74 & .65 & .71 & .76 & .68 \\
            \addlinespace
            Critique-only                   & .30 & .72 & .71 & .59 & .52 & .82 & .55\\ 
            \hspace{1em}ELA                 & .20 & .63 & .69 & .49 & .32 & .85 & .39 \\
            \hspace{1em}Math                & .41 & .79 & .74 & .65 & .68 & .77 & .67 \\
            \addlinespace
            Fusion (raw text + critique)       & .34 & .80 & .75 & .63 & .64 & .81 & .64\\
            \hspace{1em}ELA                      & .27 & .72 & .73 & .54 & .49 & .83 & .51 \\
            \hspace{1em}Math                     & .43 & .86 & .78 & .70 & .77 & .79 & .73\\
            \addlinespace
            ELA Fusion (stand-alone)       & .33 & .69 & .69 & .47 & .53 & .75 & .50\\
            Math Fusion (stand-alone)      & .39 & .85 & .79 & .72 & .73 & .82 & .73\\ 
            \bottomrule
            \end{tabular}
        \smallskip
        \tablenote{ELA = English language arts, p = proportion of items predicted as reject, AUC = area under the curve. Cutoff threshold for models were fixed to .5.}            
    \end{threeparttable}
\end{table*}

\subsection{Improving Sensitivity by Lowering the Cutoff Threshold}
\begin{figure*}[!t] 
    \caption{Cutoff threshold analysis for fusion model (raw text + critique)}
    \includegraphics[width=\textwidth]{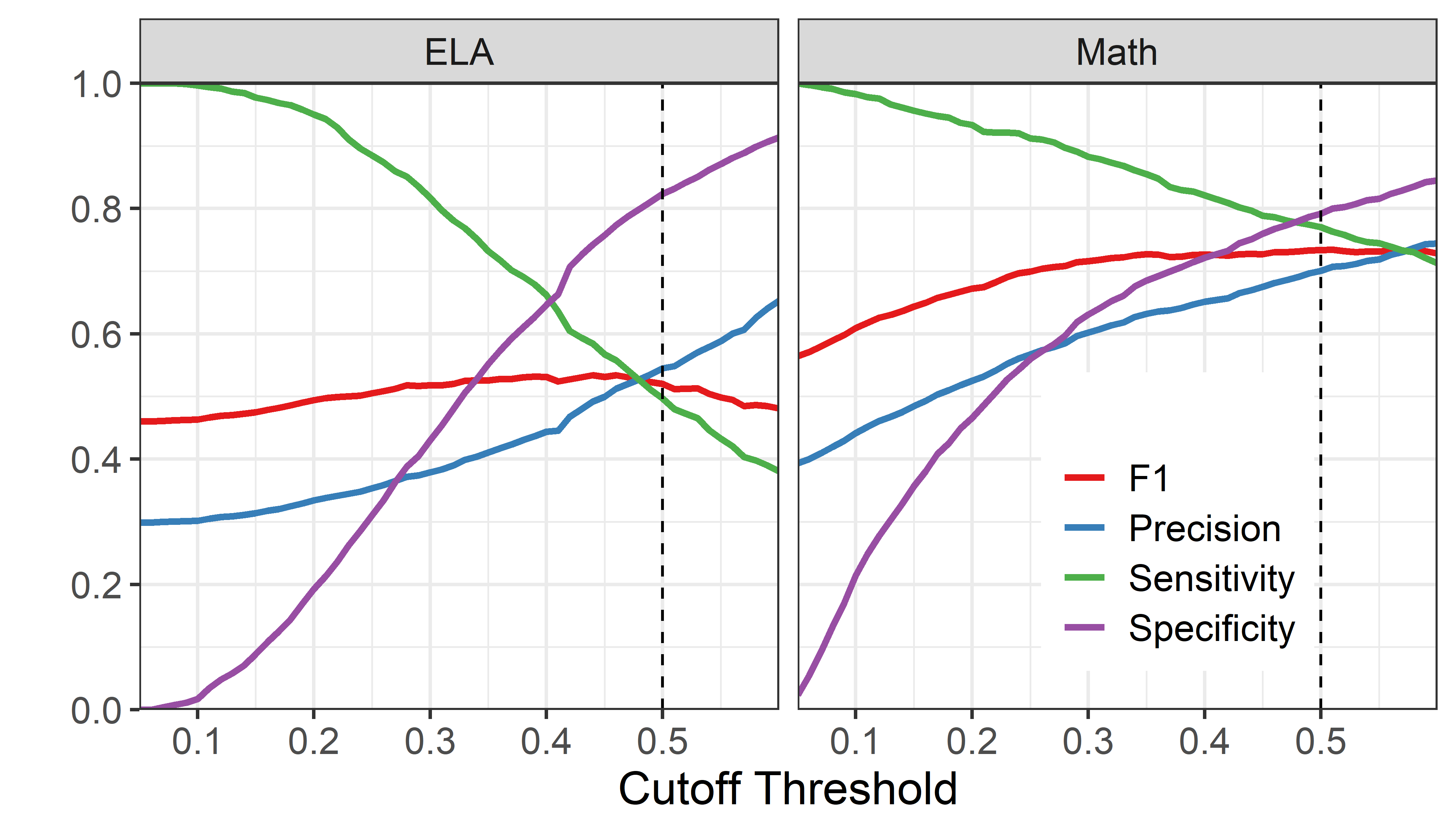}
    \figurenote{ELA = English language arts. Lowering the cutoff threshold below .5 increases sensitivity to flag rejected items at the cost of specificity. A lower cutoff threshold may be preferable in contexts such as automated item generation where item generation has a low cost. }
    \label{fig:thresh}    
\end{figure*}

Lowering the cutoff threshold below .5 increases sensitivity to flag non-viable items at the cost of specificity (see Figure~\ref{fig:thresh}). For instance, a threshold of .25 raises the overall sensitivity to .90 while reducing specificity to .42 and F1 to .60. Math would have a sensitivity of .91 (specificity = .56, F1 = .70), while ELA would have a sensitivity of .88 (specificity = .31, F1 = .51; see~\ref{sec:results25}). A low detection threshold may be preferable in AIG, where producing items is far less costly than evaluating them.

\subsection{Item Rejection Reasons}
\begin{figure*}[!t] 
    \caption{Boxplots of predicted rejection probability by rejected reason for items truly labeled as rejected.}
    \includegraphics[width=\textwidth]{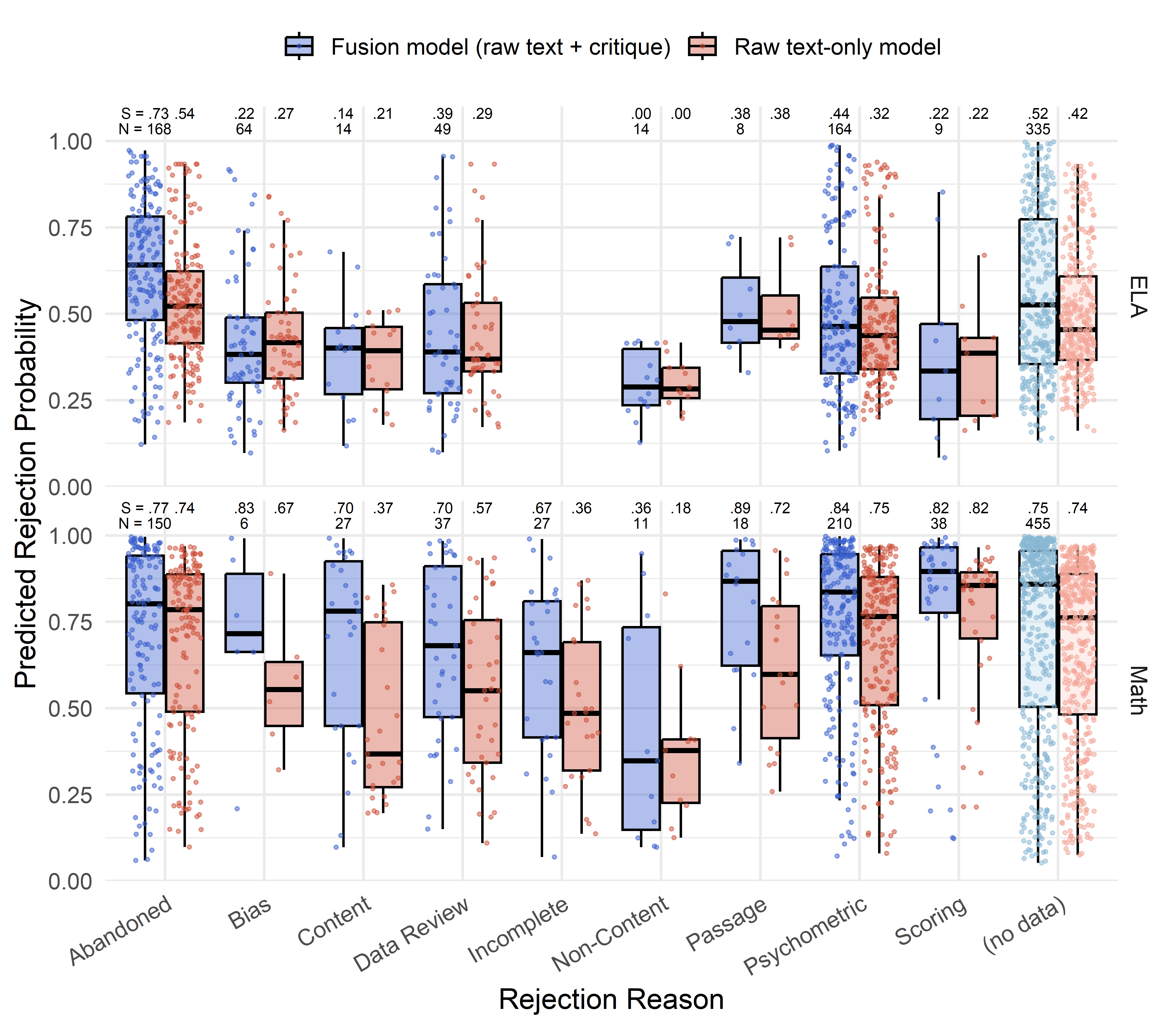} 
    \figurenote{ELA = English language arts. S=Sensitivity (using $\geq .5$ probability threshold), N=number of items, Scoring = issues with rubrics, answer keys, scoring logic, or scoring metadata, Abandoned = items without a suitable exam placement, Passage = rejection driven by the passage set rather than the item itself, Psychometric = psychometric issues like low discrimination and extreme difficulty, Incomplete = items that never reached field-testing due to stalling in development, Data review = questionable psychometric qualities requiring content expert review, including C-level differential item functioning, Content = content errors like standard misalignment, Bias = content accessibility, bias, fairness, or sensitivity concerns, Non-content = issues like missing, incompatible, or corrupt data that is not expected to be detectable by text-based models, (no data) = items rejected without comments or with unclear reasons. Only items truly labeled as rejected are shown. For full rejection reason details, see~\ref{sec:rejection_reasons}.}
    \label{fig:rejectionreason}    
\end{figure*}

Predicted rejection probability by rejection reason was estimated on truly rejected items in the test data (see Figure~\ref{fig:rejectionreason}). Items that were abandoned (sensitivity $=.75$) or had psychometric issues (sensitivity $= .66$) had one of the highest sensitivities for the fusion model, especially for math. Compared to the text-only model, incorporating item critiques in the fusion model improved detection across nearly all rejection categories. Low sensitivity (.16) to detect non-content rejections was expected as they typically reflect back-end data integrity issues unrelated to item text. Detection rate increased substantially for incomplete math items ($\Delta$ sensitivity $= .31$) and those with content issues ($\Delta$ sensitivity $= .33$). However, the fusion model struggled to detect concerns regarding bias, sensitivity, fairness, and accessibility (sensitivity $= .27$), most of which were ELA (91\%). 

\subsection{Item Difficulty and Discrimination}

\begin{figure*}[!t] 
    \caption{Item difficulty and discrimination by fusion model-predicted rejection probability.}
    \includegraphics[width=\textwidth]{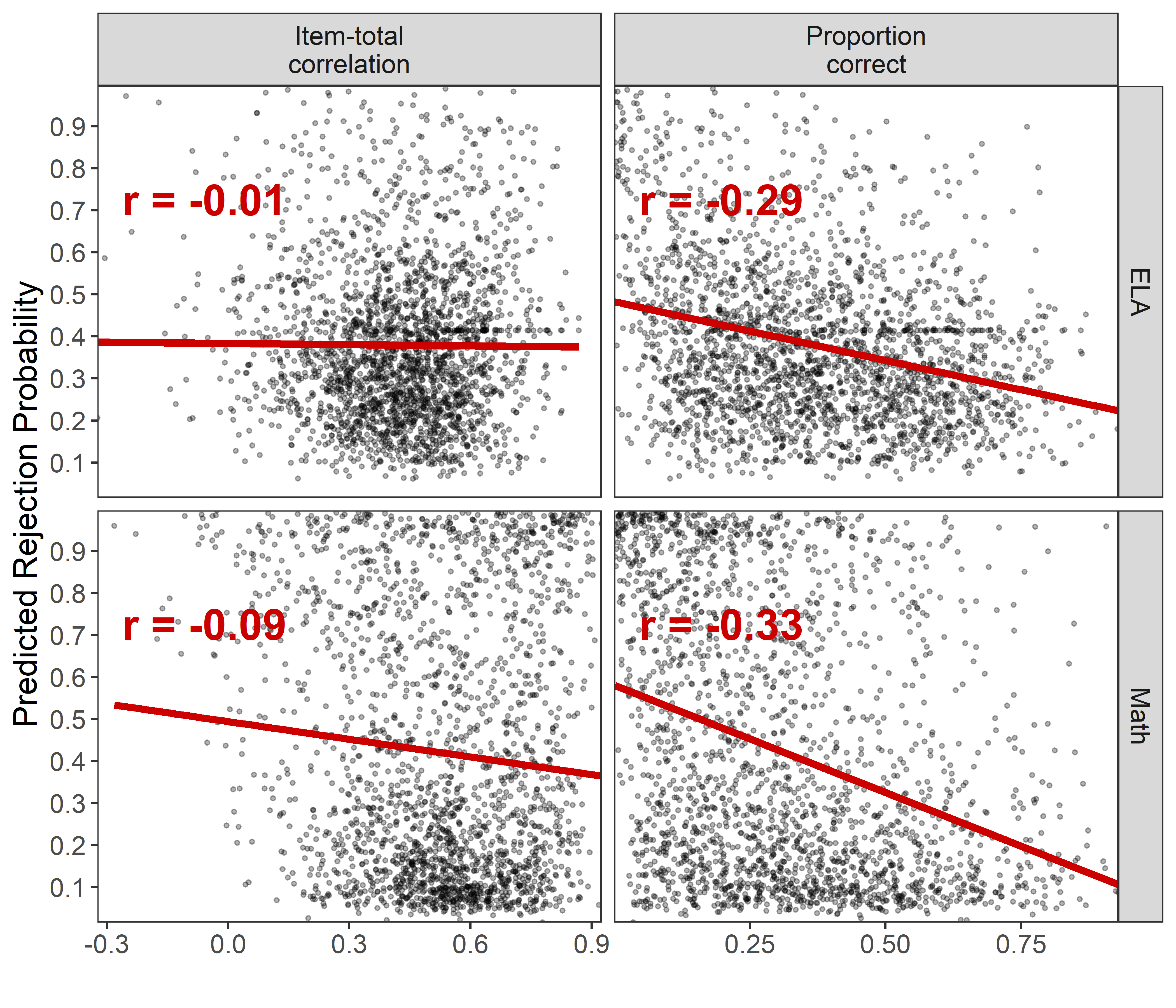} 
    \figurenote{r = Spearman correlations.}
    \label{fig:stats}    
\end{figure*}

Spearman correlations between predicted rejection probability and several psychometric statistics was strongest with proportion correct ($r= - .30$), indicating that the model assigned higher rejection probabilities to more difficult items. Correlations with total item correlation ($r = - .06$), IRT discrimination ($r = -.07$), and the item intercept parameter ($-1.7ab$; $r = - .19$) were weaker. Overall, the model is sensitive to rejecting items that may be too difficult for the target population (see Figure~\ref{fig:stats}). Note that 6.4\% of items had missing item statistics because they were rejected prior to field test analyses. For these items, sensitivity was 66\%, which was the same as the other items.

\subsection{Critique Sentiment}
Correlations between critique sentiment and $P_{text}$ and $P_{fusion}$ was significant for only math $P_{fusion}$ ($r = -.05$, $p< .05$). Negative critique sentiment was associated with slightly increased $P\Delta$, across all subject and status label groups. The Spearman correlation between sentiment and $P\Delta$ was $-.05$ for ELA ($p<.01$) and $-.14$ for math ($p<.0001$). Disaggregating by status label, correlations were not significantly different from zero for ELA items ($p>.05$), and $-.11$ for accepted and $-.18$ for rejected math items ($p<.0001$). Therefore, the association was stronger for math than for ELA, and strongest among math items that had a true rejected status. Results were consistent with the interpretation that the critiques contributed information beyond the item text itself, where negative sentiment increased rejection rates. But overall, sentiment was a minor characteristic of how critiques were related to the model behavior.

\section{Discussion}

We began the study with the goal of building a comprehensive AIE model that could identify unwanted items across as many aspects of item quality as possible. We achieved this by training a transformer language model to classify items as accepted or rejected for operational use, using item status data from a large-scale standardized testing program. Data augmentation using item critiques generated by Qwen3 improved prediction accuracy across nearly all rejection reasons. The result was a single AIE model that is nearly comprehensive, much more convenient than most other AIE methods that target a single aspect of item quality. Our approach showed practical promise for large-scale assessment pipelines where both the evaluation and discard of items carry substantial cost.

We want to emphasize that AIE is useful for both human-written and AI-generated items, though the purpose and strategy of its use can vary. In AIG contexts, where item generation is inexpensive, a lower detection threshold such as .25 may be preferable. At the cost of specificity (.42) and F1 (.60), sensitivity of our fusion model rises to .90, catching most future rejections before they happen. This can save a substantial amount of time spent on content expert review. In more traditional item development contexts, the default threshold of .5 may serve as an effective early warning to revise the item. The predictions were considerably more accurate for math than ELA, which may suggest fundamental differences in how items are rejected between these subjects. Potentially, ELA item rejections were often related to their passages, which were excluded from the current study. 

Among all item statistics examined, proportion correct (item difficulty) showed the strongest relationship with rejection. In spite of not directly using item statistics in the training, the model especially tended to flag items that were too difficult. This shows the value of item difficulty prediction research \parencite{alkhuzaey2024text, benedetto2023survey}, but the modest correlation we found ($r = - .30$) suggests that difficulty prediction alone is insufficient, as items are rejected for many reasons beyond difficulty.

Several aspects of this study are novel. First, the approach of fusing an encoder representation of raw item text with an encoder representation of a separate decoder-generated critique is relatively new, though a handful of studies have taken similar approaches \parencite{henrichsen2025twostagereasoninginfusedlearningimproving, hsieh2023distillingstepbystepoutperforminglarger, scarlatos2025smart, feng2025reasoningsamplingaugmentedmcqdifficulty}. This method leverages the decoder's capacity to draw on broad world knowledge when generating critiques. Decoders are not designed to output numerical estimates reliably, so we rely on the encoder for precise, task-specific prediction. Our approach could likely improve by using a larger decoder. The Qwen3-0.6B model we used is very small (i.e., 600 million parameters) compared to the most powerful proprietary multimodal models rumored to have trillions of parameters \parencite{li2026incompressibleknowledgeprobesestimating}. The caveat is that these proprietary models may require additional security measures unlike models that can be ran locally like Qwen. The field of AIE could therefore benefit from testing more powerful decoders combined with efficient, precisely fine-tuned encoders.

Another novel aspect of this study was the use of item developer comments. We are not aware of any other paper that has used item developer comments to help build a prediction model in the educational assessment context. The closest precedent may be \cite{ma2025examinee}, who applied DistilBERT and machine learning to examinee comments collected after test administration, building a model to identify comments most relevant for item review. It established that free-text commentary contains information for item quality decisions. We considered using item developer comments in the training data but decided against it, as only 19\% of our data had rejection comments. Nevertheless, we could benefit from further exploration of how item developer comments can be leveraged for AIE.

Our model takes a nearly comprehensive approach to AIE, as the key outcome of judging item quality is whether it is acceptable for operational use. However, we cannot fully claim comprehensiveness for several reasons. For example, item quality partly depends on the other items already in the bank. We do not want two or more items that are too similar to each other, and detecting this would require a similarity analysis \parencite{peng2020automatic}. Further, the distribution of item difficulty within a bank or test should typically be balanced \parencite{vanderLinden2000}. Evaluating these require comparing the new items to the current item bank.

Finally, we acknowledge that some rejection reasons are inherently unpredictable from item content alone, including item exposure, rejection due to other items within the same passage set, data integrity issues, or dependence on future events not yet reflected in the data. Incorporating full item metadata including the passage text and scoring rubric into the prediction could partly resolve this, but some item quality issues will always depend on unpredictable future events.

One of the biggest concerns with our fusion model was its difficulty identifying items flagged for bias, sensitivity, fairness, or accessibility concerns, possibly because these judgments require contextual and cultural knowledge that is difficult to encode from item text alone. This suggests that human review remains essential for detecting these issues in particular.

\section{Conclusion}
We introduced a novel approach to automated item evaluation (AIE) by training a transformer language model to predict item acceptance or rejection using item status data from a large-scale standardized testing program. Augmenting item text with LLM-generated critiques improved prediction accuracy across nearly all rejection reasons, demonstrating that combining encoder and decoder representations is a promising direction for AIE. Unlike prior work that has focused on predicting individual item properties such as difficulty, our approach targets the operational decision that ultimately determines an item's fate, offering a single, near-comprehensive quality metric applicable to both human-written and AI-generated items. As AIG continues to expand the volume of items requiring review, scalable AIE methods like the one presented here offer practical promise for reducing the cost and burden of item evaluation in large-scale assessment programs.


\bibliographystyle{apalike}

\bibliography{bibliography}

@article{hartmann2023,
  author  = {Hartmann, Jochen and Heitmann, Mark and Siebert, Christian and Schamp, Christina},
  title   = {More than a Feeling: Accuracy and Application of Sentiment Analysis},
  journal = {International Journal of Research in Marketing},
  year    = {2023},
  volume  = {40},
  number  = {1},
  pages   = {75--87},
  doi     = {10.1016/j.ijresmar.2022.05.005}
}

@article{Gorgun2025,
author = {Gorgun, Guher and Bulut, Okan},
title = {Instruction-Tuned Large-Language Models for Quality Control in Automatic Item Generation: A Feasibility Study},
journal = {Educational Measurement: Issues and Practice},
volume = {44},
number = {1},
pages = {96-107},
doi = {https://doi.org/10.1111/emip.12663},
url = {https://onlinelibrary.wiley.com/doi/abs/10.1111/emip.12663},
eprint = {https://onlinelibrary.wiley.com/doi/pdf/10.1111/emip.12663},
year = {2025}
}

@Inbook{vanderLinden2000,
author="van der Linden, Wim J.
and Pashley, Peter J.",
title="Item Selection and Ability Estimation in Adaptive Testing",
bookTitle="Computerized Adaptive Testing: Theory and Practice",
year="2000",
publisher="Springer Netherlands",
address="Dordrecht",
pages="1--25",
isbn="978-0-306-47531-3",
doi="10.1007/0-306-47531-6_1",
url="https://doi.org/10.1007/0-306-47531-6_1"
}

@phdthesis{peng2020automatic,
  author    = {Peng, Fang},
  title     = {Automatic enemy item detection using natural language processing},
  school    = {University of Illinois Chicago},
  year      = {2020},
  doi       = {10.25417/uic.14134559.v1},
  url       = {https://doi.org/10.25417/uic.14134559.v1}
}

@article{li2026incompressibleknowledgeprobesestimating,
      title={Incompressible Knowledge Probes: Estimating Black-Box {LLM} Parameter Counts via Factual Capacity}, 
      author={Bojie Li},
      year={2026},
      journal={arXiv preprint arXiv:2604.24827},
      archivePrefix={arXiv},
      primaryClass={cs.LG},
      url={https://arxiv.org/abs/2604.24827}, 
}

@misc{amini2025promptingstrategieslanguagemodelbased,
      title={Prompting Strategies for Language Model-Based Item Generation in K-12 Education: Bridging the Gap Between Small and Large Language Models}, 
      author={Mohammad Amini and Babak Ahmadi and Xiaomeng Xiong and Yilin Zhang and Christopher Qiao},
      year={2025},
      eprint={2508.20217},
      archivePrefix={arXiv},
      primaryClass={cs.CL},
      url={https://arxiv.org/abs/2508.20217}, 
}

@article{pelanek2022towards,
  title={Towards design-loop adaptivity: identifying items for revision},
  author={Pel{\'a}nek, Radek and Effenberger, Tom{\'a}{\v{s}} and Kuku{\v{c}}ka, Adam and others},
  journal={Journal of Educational Data Mining},
  volume={14},
  number={3},
  pages={1--25},
  year={2022}
}

@incollection{shin2024automated,
  title={Automated short-response scoring for automated item generation in science assessments},
  author={Shin, Jinnie and Gierl, Mark J},
  booktitle={The Routledge International Handbook of Automated Essay Evaluation},
  pages={504--534},
  year={2024},
  publisher={Routledge}
}

@inproceedings{wang2025using,
  title={Using generated rubrics to provide a window into item evaluation with multi-agent {LLMs}},
  author={Wang, Yu and Gopalakrishnan, Madhumitha and Bergner, Yoav},
  booktitle={International Conference on Artificial Intelligence in Education},
  pages={203--217},
  year={2025},
  organization={Springer}
}

@incollection{yuaie,
    author = {Yu, Martin C and Burke, Maura I},
    editor = {Thompson, Isaac and Yankov, Georgi P and Hernandez, Ivan},
    isbn = {9780197807279},
    title = {Automatic Item Generation, Evaluation, and Scale Construction of Non-Cognitive Measures with Generative Language Models},
    booktitle = {Artificial Intelligence for I-O Psychologists: Research and Applications},
    publisher = {Oxford University Press},
    year = {2026},
    month = {06},
    doi = {10.1093/9780197807309.003.0010},
    url = {https://doi.org/10.1093/9780197807309.003.0010},
    eprint = {https://academic.oup.com/book/0/chapter/561205002/chapter-pdf/68115481/isbn-9780197807309-book-part-10.pdf},
}

@Article{electronics14112279,
AUTHOR = {Prentzas, Jim and Binopoulou, Ariadni},
TITLE = {Explainable Artificial Intelligence Approaches in Primary Education: A Review},
JOURNAL = {Electronics},
VOLUME = {14},
YEAR = {2025},
NUMBER = {11},
ARTICLE-NUMBER = {2279},
URL = {https://www.mdpi.com/2079-9292/14/11/2279},
ISSN = {2079-9292},
DOI = {10.3390/electronics14112279}
}

@article{oluoke_inpress,
  author    = {Oluoke, Omopekunola Moses and Gorbacheva, Anna M. and Monday, Onoja Abah},
  title     = {Evaluating {LLM}-Generated Assessment Items: A {JBI}-Guided Critical Appraisal and Checklist},
  journal   = {International Journal of Evaluation and Research in Education},
  year      = {2026},
  note      = {In press}
}

@article{tan2025review,
  title={A review of automatic item generation techniques leveraging large language models},
  author={Tan, Bin and Armoush, Nour and Mazzullo, Elisabetta and Bulut, Okan and Gierl, Mark},
  journal={International Journal of Assessment Tools in Education},
  volume={12},
  number={2},
  pages={317--340},
  year={2025},
  publisher={{\.I}zzet KARA}
}

@article{falcao2022feasibility,
  title={Feasibility assurance: a review of automatic item generation in medical assessment},
  author={Falc{\~a}o, Filipe and Costa, Patr{\'\i}cio and P{\^e}go, Jos{\'e} M},
  journal={Advances in Health Sciences Education},
  volume={27},
  number={2},
  pages={405--425},
  year={2022},
  publisher={Springer}
}

@ARTICLE{circi2023automatic,
    
AUTHOR={Circi, Ruhan  and Hicks, Juanita  and Sikali, Emmanuel },
           
TITLE={Automatic item generation: foundations and machine learning-based approaches for assessments},
          
JOURNAL={Frontiers in Education},
          
VOLUME={Volume 8 - 2023},
  
YEAR={2023},
  
URL={https://www.frontiersin.org/journals/education/articles/10.3389/feduc.2023.858273},
  
DOI={10.3389/feduc.2023.858273},
  
ISSN={2504-284X}}

@article{openai2024gpt4ocard,
      title={{GPT}-4o System Card}, 
      author={OpenAI},
      year={2024},
      journal={arXiv preprint arXiv:2410.21276},
      archivePrefix={arXiv},
      primaryClass={cs.CL},
      url={https://arxiv.org/abs/2410.21276}, 
}

@article{hsieh2023distillingstepbystepoutperforminglarger,
      title={Distilling Step-by-Step! Outperforming Larger Language Models with Less Training Data and Smaller Model Sizes}, 
      author={Cheng-Yu Hsieh and Chun-Liang Li and Chih-Kuan Yeh and Hootan Nakhost and Yasuhisa Fujii and Alexander Ratner and Ranjay Krishna and Chen-Yu Lee and Tomas Pfister},
      year={2023},
      journal={arXiv preprint arXiv:2305.02301},
      archivePrefix={arXiv},
      primaryClass={cs.CL},
      url={https://arxiv.org/abs/2305.02301}, 
}

@article{henrichsen2025twostagereasoninginfusedlearningimproving,
      title={Two-Stage Reasoning-Infused Learning: Improving Classification with {LLM}-Generated Reasoning}, 
      author={Mads Henrichsen and Rasmus Krebs},
      year={2025},
      journal={arXiv preprint arXiv:2507.00214},
      archivePrefix={arXiv},
      primaryClass={cs.CL},
      url={https://arxiv.org/abs/2507.00214}, 
}

@article{feng2025reasoningsamplingaugmentedmcqdifficulty,
      title={Reasoning and Sampling-Augmented MCQ Difficulty Prediction via {LLMs}}, 
      author={Wanyong Feng and Peter Tran and Stephen Sireci and Andrew Lan},
      year={2025},
      journal={arXiv preprint arXiv:2503.08551},
      archivePrefix={arXiv},
      primaryClass={cs.AI},
      url={https://arxiv.org/abs/2503.08551}, 
}

@misc{anthropic2025claude,
  author       = {Anthropic},
  title        = {Claude {Sonnet} 4.6},
  year         = {2025},
  howpublished = {\url{https://www.anthropic.com}},
  note         = {Large language model}
}

@article{lu2025smalllanguagemodelssurvey,
      title={Small Language Models: Survey, Measurements, and Insights}, 
      author={Zhenyan Lu and Xiang Li and Dongqi Cai and Rongjie Yi and Fangming Liu and Xiwen Zhang and Nicholas D. Lane and Mengwei Xu},
      year={2025},
      journal={arXiv preprint arXiv:2409.15790},
      archivePrefix={arXiv},
      primaryClass={cs.CL},
      url={https://arxiv.org/abs/2409.15790}, 
}

@article{kolesnikova2026estimating,
  title={Estimating Item Difficulty with Large Language Models as Experts},
  author={Kolesnikova, Diana and Fedyanin, Kirill and Hofman, Abe D and Brinkhuis, Matthieu JS and Bolsinova, Maria},
  journal={arXiv preprint arXiv:2605.18562},
  year={2026}
}

@misc{fu2025textbasedapproachesitemalignment,
      title={Text-Based Approaches to Item Alignment to Content Standards in Large-Scale Reading \& Writing Tests}, 
      author={Yanbin Fu and Hong Jiao and Tianyi Zhou and Nan Zhang and Ming Li and Qingshu Xu and Sydney Peters and Robert W. Lissitz},
      year={2025},
      journal={arXiv preprint arXiv:2509.26431},
      archivePrefix={arXiv},
      primaryClass={cs.CL},
      url={https://arxiv.org/abs/2509.26431}, 
}

@article{haladyna1989taxonomy,
  author    = {Thomas M. Haladyna and Steven M. Downing},
  title     = {Taxonomy of Multiple-Choice Item-Writing Rules},
  journal   = {Applied Measurement in Education},
  volume    = {2},
  number    = {1},
  pages     = {37--50},
  year      = {1989},
  doi       = {10.1207/s15324818ame0201_3}
}

@article{multimodal2025,
  author  = {Maeda, Hotaka and Lu, Yikai},
  title   = {Multimodal Test Item Parameter Prediction from Text, Images, and Metadata: Fusing Together {AI} Vision and Language Models},
  journal = {Educational and Psychological Measurement},
  year    = {2026},
  doi     = {10.1177/00131644261460779},
  url     = {https://doi.org/10.1177/00131644261460779}
}

@inproceedings{yaneva-etal-2020-predicting,
    title = "Predicting Item Survival for Multiple Choice Questions in a High-Stakes Medical Exam",
    author = "Yaneva, Victoria  and
      Ha, Le An  and
      Baldwin, Peter  and
      Mee, Janet",
    editor = "Calzolari, Nicoletta  and
      B{\'e}chet, Fr{\'e}d{\'e}ric  and
      Blache, Philippe  and
      Choukri, Khalid  and
      Cieri, Christopher  and
      Declerck, Thierry  and
      Goggi, Sara  and
      Isahara, Hitoshi  and
      Maegaard, Bente  and
      Mariani, Joseph  and
      Mazo, H{\'e}l{\`e}ne  and
      Moreno, Asuncion  and
      Odijk, Jan  and
      Piperidis, Stelios",
    booktitle = "Proceedings of the Twelfth Language Resources and Evaluation Conference",
    month = may,
    year = "2020",
    address = "Marseille, France",
    publisher = "European Language Resources Association",
    url = "https://aclanthology.org/2020.lrec-1.841/",
    pages = "6812--6818",
    language = "eng",
    ISBN = "979-10-95546-34-4", 
}

@article{ma2025examinee,
  author    = {Ye Ma},
  title     = {Analyzing Examinee Comments using {DistilBERT} and Machine Learning to Ensure Quality Control in Exam Content},
  journal   = {arXiv preprint arXiv:2504.06465},
  year      = {2025},
  url       = {https://arxiv.org/abs/2504.06465}
}

@article{qwen3technicalreport,
      title={Qwen3 Technical Report}, 
      author={{Qwen Team}},
      year={2025},
      journal={arXiv preprint arXiv:2505.09388},
      archivePrefix={arXiv},
      primaryClass={cs.CL},
      url={https://arxiv.org/abs/2505.09388}, 
}

@article{muraki1992generalized,
  title={A generalized partial credit model: Application of an EM algorithm},
  author={Muraki, Eiji},
  journal={ETS Research Report Series},
  volume={1992},
  number={1},
  pages={i--30},
  year={1992},
  publisher={Wiley Online Library}
}

@inproceedings{scarlatos2025smart,
  title={Smart: Simulated students aligned with item response theory for question difficulty prediction},
  author={Scarlatos, Alexander and Fernandez, Nigel and Ormerod, Christopher and Lottridge, Susan and Lan, Andrew},
  booktitle={Proceedings of the 2025 Conference on Empirical Methods in Natural Language Processing},
  pages={25082--25105},
  year={2025}
}

@inproceedings{benedetto2020r2de,
  title={R2DE: a NLP approach to estimating IRT parameters of newly generated questions},
  author={Benedetto, Luca and Cappelli, Andrea and Turrin, Roberto and Cremonesi, Paolo},
  booktitle={Proceedings of the tenth international conference on learning analytics \& knowledge},
  pages={412--421},
  year={2020}
}

@article{liu2019roberta,
  title={Roberta: A robustly optimized bert pretraining approach},
  author={Liu, Yinhan and Ott, Myle and Goyal, Naman and Du, Jingfei and Joshi, Mandar and Chen, Danqi and Levy, Omer and Lewis, Mike and Zettlemoyer, Luke and Stoyanov, Veselin},
  journal={arXiv preprint arXiv:1907.11692},
  year={2019}
}

@article{wauters2012item,
  title={Item difficulty estimation: An auspicious collaboration between data and judgment},
  author={Wauters, Kelly and Desmet, Piet and Van Den Noortgate, Wim},
  journal={Computers \& Education},
  volume={58},
  number={4},
  pages={1183--1193},
  year={2012},
  publisher={Elsevier}
}

@article{hsu2018automated,
  title={Automated estimation of item difficulty for multiple-choice tests: An application of word embedding techniques},
  author={Hsu, Fu-Yuan and Lee, Hahn-Ming and Chang, Tao-Hsing and Sung, Yao-Ting},
  journal={Information Processing \& Management},
  volume={54},
  number={6},
  pages={969--984},
  year={2018},
  publisher={Elsevier}
}

@article{li2025item,
  title={Item difficulty modeling using fine-tuned small and large language models},
  author={Li, Ming and Jiao, Hong and Zhou, Tianyi and Zhang, Nan and Peters, Sydney and Lissitz, Robert W},
  journal={Educational and Psychological Measurement},
  volume={85},
  number={6},
  pages={1065--1090},
  year={2025},
  publisher={SAGE Publications Sage CA: Los Angeles, CA}
}

@article{hu2022lora,
  title={Lora: Low-rank adaptation of large language models.},
  author={Hu, Edward J and Shen, Yelong and Wallis, Phillip and Allen-Zhu, Zeyuan and Li, Yuanzhi and Wang, Shean and Wang, Lu and Chen, Weizhu and others},
  journal={ICLR},
  volume={1},
  number={2},
  pages={3},
  year={2022}
}

@article{benedetto2023survey,
  title={A survey on recent approaches to question difficulty estimation from text},
  author={Benedetto, Luca and Cremonesi, Paolo and Caines, Andrew and Buttery, Paula and Cappelli, Andrea and Giussani, Andrea and Turrin, Roberto},
  journal={ACM Computing Surveys},
  volume={55},
  number={9},
  pages={1--37},
  year={2023},
  publisher={ACM New York, NY}
}

@article{alkhuzaey2024text,
  title={Text-based question difficulty prediction: A systematic review of automatic approaches},
  author={AlKhuzaey, Samah and Grasso, Floriana and Payne, Terry R and Tamma, Valentina},
  journal={International Journal of Artificial Intelligence in Education},
  volume={34},
  number={3},
  pages={862--914},
  year={2024},
  publisher={Springer}
}

@inproceedings{han2025leveraging,
  title={Leveraging Fine-tuned Large Language Models in Item Parameter Prediction},
  author={Han, Suhwa and Rijmen, Frank and Boykin, Allison Ames and Lottridge, Susan},
  booktitle={Proceedings of the Artificial Intelligence in Measurement and Education Conference (AIME-Con): Full Papers},
  pages={250--264},
  year={2025}
}

@incollection{birnbaum1968latent,
  author    = {Birnbaum, A.},
  title     = {Some latent trait models and their use in inferring an examinee’s ability},
  booktitle = {Statistical theories of mental test scores},
  editor    = {Lord, F. M. and Novick, M. R.},
  pages     = {397--479},
  publisher = {Addison-Wesley},
  year      = {1968}
}

@article{alkhuzaey2023text,
  title={Text-based question difficulty prediction: A systematic review of automatic approaches},
  author={AlKhuzaey, Samah and Grasso, Floriana and Payne, Terry R and Tamma, Valentina},
  journal={International Journal of Artificial Intelligence in Education},
  pages={1--53},
  year={2023},
  publisher={Springer}
}

@article{he2021debertav3improvingdebertausing,
      title={DeBERTaV3: Improving DeBERTa using ELECTRA-Style Pre-Training with Gradient-Disentangled Embedding Sharing}, 
      author={Pengcheng He and Jianfeng Gao and Weizhu Chen},
      year={2021},
      journal={arXiv preprint arXiv:2111.09543},
      archivePrefix={arXiv},
      primaryClass={cs.CL},
      url={https://arxiv.org/abs/2111.09543}, 
}

@article{vaswani2017attentionneed,
      title={Attention Is All You Need}, 
      author={Ashish Vaswani and Noam Shazeer and Niki Parmar and Jakob Uszkoreit and Llion Jones and Aidan N. Gomez and Lukasz Kaiser and Illia Polosukhin},
      year={2017},
      journal={arXiv preprint arXiv:1706.03762},
      archivePrefix={arXiv},
      primaryClass={cs.CL},
      url={https://arxiv.org/abs/1706.03762}, 
}

@inproceedings{Devlin2019,
   author = {Jacob Devlin and Ming Wei Chang and Kenton Lee and Kristina Toutanova},
   booktitle = {NAACL HLT 2019 - 2019 Conference of the North American Chapter of the Association for Computational Linguistics: Human Language Technologies - Proceedings of the Conference},
   title = {{BERT}: Pre-training of deep bidirectional transformers for language understanding},
   volume = {1},
   year = {2019},
}

@Book{standards,
  title = {{Standards for Educational and Psychological Testing}},
  author = {{AERA} and {APA} and {NCME}},
  publisher = {American Educational Research Association},
  address = {Washington, DC},
  year = {2014}
}

@article{maeda2024field,
author = {Hotaka Maeda},
title ={Field-Testing Multiple-Choice Questions With {AI} Examinees: English Grammar Items},
journal = {Educational and Psychological Measurement},
volume = {85},
number = {2},
pages = {221-244},
year = {2025},
doi = {10.1177/00131644241281053},
URL = {https://doi.org/10.1177/00131644241281053},
eprint = {https://doi.org/10.1177/00131644241281053}
}

@article{warner2024smarter,
      title={Smarter, Better, Faster, Longer: A Modern Bidirectional Encoder for Fast, Memory Efficient, and Long Context Finetuning and Inference}, 
      author={Benjamin Warner and Antoine Chaffin and Benjamin Clavié and Orion Weller and Oskar Hallström and Said Taghadouini and Alexis Gallagher and Raja Biswas and Faisal Ladhak and Tom Aarsen and Nathan Cooper and Griffin Adams and Jeremy Howard and Iacopo Poli},
      year={2024},
      journal={arXiv preprint arXiv:2412.13663},
      archivePrefix={arXiv},
      primaryClass={cs.CL},
      url={https://arxiv.org/abs/2412.13663}, 
}

@article{maeda_finding_2025,
	title ={Finding words associated with {DIF}: Predicting differential item functioning using {LLMs} and explainable {AI}},
	volume = {62},
	issn = {1745-3984},
	shorttitle = {Finding {Words} {Associated} with {DIF}},
	url = {https://onlinelibrary.wiley.com/doi/abs/10.1111/jedm.70017},
	doi = {10.1111/jedm.70017},
	language = {en},
	number = {4},
	urldate = {2026-02-10},
	journal = {Journal of Educational Measurement},
	author = {Maeda, Hotaka and Lu, Yikai},
	year = {2025},
	pages = {883--906},
}

\onecolumn
\appendix 
\renewcommand{\thesection}{Appendix \Alph{section}}

\newpage
\section{Rejection Reason Definitions}
\label{sec:rejection_reasons}
\begin{enumerate}
    \item Content: content issue or error, including but not limited to (1) misalignment with the blueprint or standards, (2) unrealistic, inconsistent, or confusing scenarios, or (3) basic issues with the text or image such as grammar errors or missing information. A large portion of these items are rejected at a content review event just prior to field testing. 

    \item Psychometric: The following statistical criteria result in immediate rejection without manual review: proportion correct below .03, item-total correlation at or below .05, a distractor with a higher item-total correlation than the correct answer, poor inter-rater reliability on hand-scored items (exact agreement less than 75\%, 65\%, or 55\% on items with 2, 3, or greater score points, respectively), or polytomous items where mean estimated $\theta$ does not increase monotonically across score levels. Items can also be rejected for less severe psychometric issues, though these require review and situational judgment: low or negative discrimination, too easy, too difficulty, insufficient sample size, overexposure, parameter drift, or differential item functioning.

    \item Data review: These are items with questionable psychometric qualities that needed to be reviewed by content experts. Items flagged with C-level differential item functioning are always reviewed by content experts.

    \item Bias: Content bias, sensitivity, fairness, or accessibility reasons

    \item Abandoned: Items that could not find a suitable use in an exam. Many tend to be extra items in passage sets. Sometimes, the item itself has no content issues. 

    \item Incomplete: Items stuck in the item development process. Never reached field-testing.

    \item Passage: The entire passage set (i.e., stimulus or testlet) was rejected. Sometimes, the item itself has no content issues. 
    
    \item Scoring: Any issue related to scoring rubric, answer key, scoring logic, or scoring meta data. This can be a combination of content, data integrity, and psychometric issues.

    \item Non-content: Issues unrelated to the text or images of the item itself, such as missing metadata, corrupt or malformed data, format incompatibilities, import errors, or expired copyright permissions. Because these are primarily data integrity issues, text-based models are not expected to detect them.
    
    \item (no data): Items rejected without comments or unclear reasons. 
 \end{enumerate}

\newpage
\section{Qwen3 Zero-Shot Classification Prompt}
\label{sec:zero_prompt}
Below is an example python script for generating the Qwen3-0.6B zero-shot classification prompt. 
\begin{lstlisting}[language=Python]
SYSTEM_PROMPT = (
    "You are an expert assessment item reviewer. You will be given an assessment item. Respond with exactly one word: 'accept' or 'reject'. Respond 'accept' if the item is usable for assessment as-is. Respond 'reject' if the item has issues that go beyond a simple fix, such as: fundamental ambiguity, significant bias or sensitivity concerns, clear misalignment to the stated standard, or broken/unfixable formatting. Do not include any explanation, punctuation, or additional text."
)
def make_messages(row):
    return {
        "messages": [
            {"role": "system", "content": SYSTEM_PROMPT},
            {"role": "user", "content": f"This is a {row['subject']} subject, grade {row['grade']} item: \n{row['text']} \n{row['standards']} \nAny relevant images, figures, or tables have been excluded."
                }]}
\end{lstlisting}

\newpage
\section{Qwen3 Item Critique Prompt}
\label{sec:critique_prompt}
Below is an example python script for generating the Qwen3-0.6B item critique prompt. 
\begin{lstlisting}[language=Python]
SYSTEM_PROMPT = (
    "You are an expert assessment item reviewer with deep knowledge of psychometrics and item development best practices. You will be given an assessment item. Review the item for common quality issues such as: ambiguity, item difficulty, cultural or demographic bias and sensitivity, content alignment to standards, and flawed grammar or formatting. Write a concise 2 sentence summary: if the item has issues, name them plainly and briefly explain why they matter; if the item looks sound, say so and note its strongest quality. Do not use bullet points or headers. Be direct and specific — avoid vague praise or vague criticism."
)
def make_messages(row):
  return {
    "messages": [
            {"role": "system", "content": SYSTEM_PROMPT},
            {"role": "user", "content": f"This is a {row['subject']} subject, grade {row['grade']} item: \n{row['text']} \n{row['standards']}  \nAny relevant images and figures or tables have been excluded. \nWrite a concise 2 sentence summary about the item quality, including but not limited to its item difficulty, bias and sensitivity, and content alignment to standards."
            }]}
\end{lstlisting}

\newpage
\section{Example Qwen3 Critiques}
\label{sec:example_critiques}

\begin{table}[h]
\centering
\caption{\label{tab:critiques}Example Critiques Generated using Qwen3-0.6B.}
\footnotesize{
    \resizebox{\ifdim\width>\linewidth\linewidth\else\width\fi}{!}{
     \begin{tabular}[t]{llcp{10cm}ccc}
    \toprule
    Subject & Status & Grade & Critique  & Sentiment & $P_{fusion}$ & $P\Delta$\\ 
    \midrule
        ELA & Accept & 3 & The item presents a grade 3 ELA question about identifying similarities between Bear and Max as cats, using plausible answers from the provided choices. It appears to be a standard reading comprehension task, with clear instructions and aligned content to educational standards. No ambiguities, biases, or content misalignment are evident. & 6.8 & .77 & .44\\
     & Accept & 7 & The item presents a clear conflict between the percentage-based data (Source 1) and the generalization (Source 2), creating ambiguity. While the standards are met, the item lacks contextual clarity on how the sources' findings interact, making it a flawed structure. & -6.0 & .19 & -.03\\
     & Reject & 4 & The item contains ambiguity in the phrase "wide," leaving unclear alternatives for describing the wide chair. The correct response specifies "bumpy" instead, aligning with standards for precise language use. The item meets criteria for clarity and standard alignment. & 6.6 & .26 & .02\\
     & Reject & 4 & The item presents a question about the narrator's feelings toward being a seeing eye dog, but lacks clarity in the answer choices, leaving ambiguity about the intended response. It also includes an image/figure exclusion, which may impact the assessment's validity. No standard content alignment was identified. & -6.9 & .80 & .44\\
    Math & Accept & 6 & The item lacks necessary visual elements (excluded images/tables) and fails to meet the standard "Reporting the number of observations," as the task requires entering a count instead of identifying the number of data points. It appears to align well with the specified standards but may require additional contextual clarity to ensure correctness. & 5.2 & .10 & -.03\\ 
     & Accept & 7 & The item presents a clear mathematical statement about how multiplication of coefficients relates to variables, making it suitable for grade 7 students. However, the phrasing "supports" or "does not support" lacks specificity, leaving ambiguity around the intended meaning. The question aligns with standard algebraic reasoning but requires further clarification to ensure accuracy. The item shows strong alignment with grade-level curriculum standards. & 6.7 & .80 & .33\\
     & Reject & 7 & The item contains ambiguity in the phrasing, making it unclear how the answer hinges on x's relationship to the context. The difficulty is moderate, aligning well with grade-level standards. The content is aligned to mathematical benchmarks without bias. & 6.3 & .99 & .04\\
     & Reject & 8 & The item contains a formatting error in the equation, specifically a missing fraction sign, which affects the interpretation of the algebraic manipulation. The answer choices provide conflicting explanations regarding the equation's solution, but the correct response is that the equation has exactly one valid solution, x = 2, which aligns with standard algebraic techniques. The item lacks clarity in presenting the process, potentially leading to confusion about the expected outcome. & -6.9 & .92 & .33\\
    \bottomrule    
    \end{tabular}}
}
\tablenote{Status = true status label, Sentiment = sentiment logit (positive = positive sentiment, negative = negative sentiment), $P_{fusion}$ = fusion model probability of rejection, $P\Delta$ = $P_{fusion}$ minus text-only model probability of rejection, ELA = English language arts. Two critiques from accepted and rejected ELA and math items were randomly selected, with the requirement that one had a near-zero $P\Delta$, and another had $P\Delta > .3$.}
\end{table}

\newpage
\section{Classification with .25 Cutoff}
\label{sec:results25}

\begin{table}[h]
    \centering
    \begin{threeparttable}
        \caption{\label{tab:model_comparison25}Classification Performance by Model with .25 Cutoff Threshold} 
        \footnotesize
        \begin{tabular}[t]{lccccccccc}
            \toprule
            Model & p & AUC & Accuracy & Precision & Sensitivity & Specificity & F1\\
            \midrule
            Raw text-only & .78 & .76 & .51 & .41 & .92 & .30 & .56\\
            \hspace{1em}ELA & .84 & .69 & .42 & .33 & .94 & .20 & .49\\
            \hspace{1em}Math & .71 & .80 & .61 & .50 & .91 & .42 & .64\\
            \addlinespace
            Critique-only & .80 & .73 & .48 & .39 & .92 & .26 & .55\\
            \hspace{1em}ELA & .92 & .63 & .35 & .31 & .96 & .09 & .47\\
            \hspace{1em}Math & .67 & .79 & .62 & .50 & .88 & .46 & .64\\
            \addlinespace
            Fusion (raw text + critique) & .69 & .80 & .58 & .45 & .90 & .42 & .60\\
            \hspace{1em}ELA & .75 & .72 & .48 & .35 & .88 & .31 & .51\\
            \hspace{1em}Math & .62 & .86 & .70 & .57 & .91 & .56 & .70\\
            \addlinespace
            ELA Fusion (stand-alone) & .89 & .70 & .38 & .32 & .96 & .13 & .48\\
            Math Fusion (stand-alone) & .59 & .85 & .71 & .58 & .88 & .59 & .70\\
            \bottomrule 
            \end{tabular}
            \smallskip
        \tablenote{ELA = English language arts, p = proportion of items predicted as reject, AUC = area under the curve. Cutoff threshold for models were fixed to .25.}            
    \end{threeparttable}
\end{table}

\end{document}